\documentclass[letterpaper, 10 pt, conference]{ieeeconf}  
\usepackage{cite}
\usepackage{amsmath,amssymb,amsfonts}
\usepackage{algpseudocode}
\usepackage{graphicx}
\usepackage{textcomp}
\usepackage{xcolor}
\usepackage{amsmath,amssymb,amsfonts}
\usepackage{mathtools}
\usepackage{mathrsfs}
\usepackage{graphicx}
\usepackage{textcomp}
\usepackage{xcolor}
\usepackage{tikz}
\usepackage{units}
\usepackage{cprotect}
\usepackage{amsfonts}
\usepackage{fvextra}
\usepackage{textcomp}
\usepackage{paracol}
\usepackage{amssymb}
\usepackage{cite}
\usepackage{amsmath,amssymb,amsfonts}
\usepackage{mathtools}
\usepackage{amsmath}
\usepackage{soul}
\usepackage{graphicx}
\usepackage{textcomp}
\usepackage{xcolor}
\usepackage{tikz}
\usepackage{units}
\usepackage{tikz}
\usepackage{bm}
\usepackage{ragged2e} 
\usepackage{subcaption}
\usepackage[utf8]{inputenc}
\usepackage{wrapfig}   
\usepackage{algorithm}
\usepackage{algpseudocode}
\usepackage{hyperref}
\usepackage{makecell}
\usepackage [english]{babel}
\usepackage [autostyle, english = american]{csquotes}
\usepackage{lipsum,graphicx,multicol,subcaption}
\usepackage [english]{babel}
\usepackage[justification=centering]{caption}
\usepackage{lipsum,graphicx,multicol}
\usepackage{atbegshi}
\def\BibTeX{{\rm B\kern-.05em{\sc i\kern-.025em b}\kern-.08em
    T\kern-.1667em\lower.7ex\hbox{E}\kern-.125emX}}

\IEEEoverridecommandlockouts                              

\title{\LARGE \bf
How IMU Drift Influences Multi-Radar Inertial Odometry for Ground Robots in Subterranean Terrains
\thanks{This work has been partially funded by the European Union’s Horizon Europe Research and Innovation Programme, under the Grant Agreement No.101138451 PERSEPHONE, and in part by the Swedish Department of Energy and LKAB through the Sustainable Underground Mining (SUM) Academy Programme project ”Autonomous Drones for Underground Mining Operations”.
}
}
\author{Moumita Mukherjee, Magnus Norén, Anton Koval, Avijit Banerjee, George Nikolakopoulos}

\begin{document}
\maketitle
\thispagestyle{empty}
\pagestyle{empty}
\begin{abstract}
Reliable radar inertial odometry (RIO) requires mitigating IMU bias drift, a challenge that intensifies in subterranean environments due to extreme temperatures and gravity-induced accelerations.  Cost-effective IMUs such as the Pixhawk, when paired with FMCW TI IWR6843AOP EVM radars, suffer from drift-induced degradation compounded by sparse, noisy, and flickering radar returns, making fusion less stable than LiDAR-based odometry. Yet, LiDAR fails under smoke, dust and aerosols, whereas FMCW radars remain compact, lightweight, cost-effective, and robust to these situations. To address these challenges, we propose a two-stage MRIO framework that combines an IMU bias estimator for resilient localization and mapping in GPS-denied subterranean environments affected by smoke. In this, radar-based ego-velocity estimation is formulated through a least-squares approach and incorporated into an EKF for online IMU bias correction, thus, the corrected IMU accelerations are fused with heterogeneous measurements from multiple radars and IMU to refine odometry. The proposed framework further supports radar-only mapping by exploiting the robot’s estimated translational and rotational displacements. In subterranean field trials, MRIO delivers robust localization and mapping, outperforming EKF-RIO. It maintains accuracy across cost-efficient FMCW radar setups and different IMUs, with resilience on Pixhawk and using higher-grade units like VectorNav. The implementation will be provided as an open-source resource to the community:\url{https://github.com/LTU-RAI/MRIO}
\end{abstract}
\begin{figure}[h] 
\includegraphics [width=0.49\textwidth]{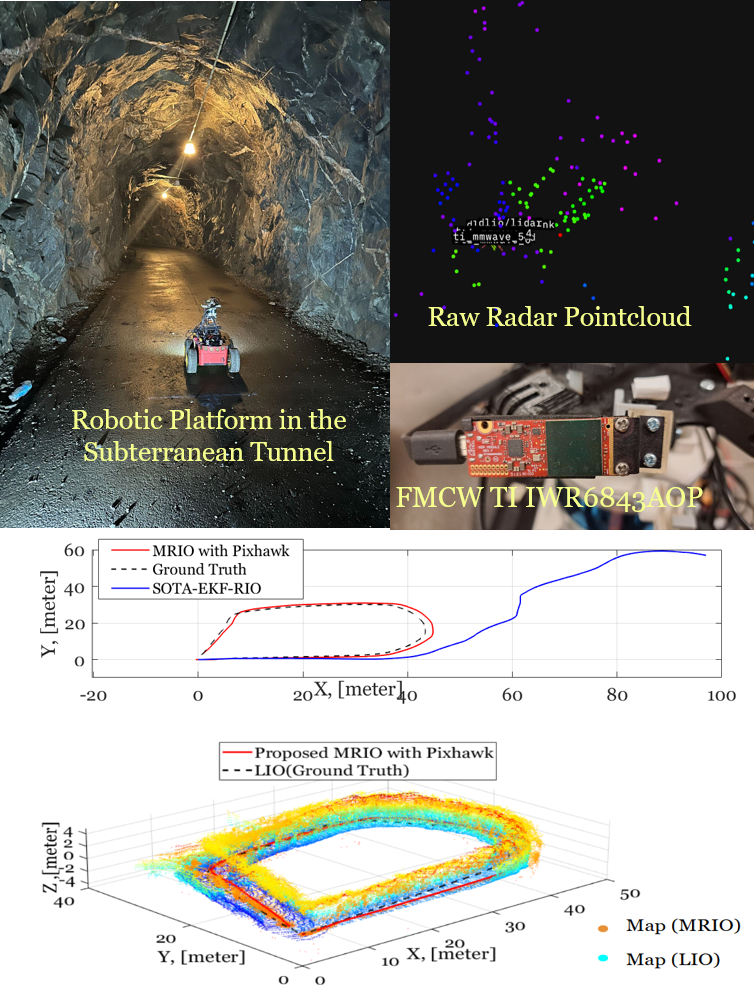}
\caption{\justifying Subterranean test environment with ground robot equipped with FMCW TI IWR6843AOP EVM radars and two IMU configurations. Experiments in a sloped, extreme-cold tunnel show the proposed MRIO framework outperforming state-of-the-art EKF-RIO with online calibration across all IMU setups.}
\label{fig:Cover_fig_1}
\end{figure}
\section{Introduction}
Following the DARPA Subterranean Challenge~\cite{tranzatto2022cerberus}, interest in autonomous navigation for subterranean (Sub-T) environments has surged, accelerating advances in perception, localization, and mapping under extreme conditions. Sub-T settings, lava tubes, mines, and tunnels are GPS-denied and low-light, impairing visual sensing. Although recent visual and LiDAR based methods enable accurate localization and mapping~\cite{taleb2025automotive}, state-of-the-art camera/LiDAR systems degrade under visually degrading conditions (VDCs), smoke, dust, and adverse weather, where visibility drops and aerosols corrupt LiDAR returns. FMCW radars~\cite{rao2017introduction} provide a resilient alternative: compact, lightweight, cost-effective, and robust to low illumination and airborne particles; they yield point clouds with spatial coordinates $(x,y,z)$ and Doppler velocity $v_d$ for motion and environment estimation. Multi-radar setups extend coverage in subterranean, GPS-denied environments, but radar point clouds are sparse, noisy, and multipath-cluttered, necessitating inertial fusion to stabilize estimates. However, IMU performance varies with operational conditions, particularly in unstructured subterranean tunnels where uneven slopes introduce uncompensated gravity components along the robot’s heading, and in extreme cold, where the IMU is prone to dynamic bias in acceleration measurements. Consequently, the choice of IMU sensor, due to its bias and noise characteristics, significantly affects the performance of radar inertial odometry, particularly in terms of accuracy under inconsistent and sparse radar measurements.
\subsection{Related Literature}
Radar based sensing has emerged as a promising approach for navigation in challenging environments, owing to its robustness to low light, dust, smoke, and other visually degrading conditions. Radar sensors can be categorized based on the structural characteristics of the point clouds they generate. These include FMCW radars~\cite{herraez2024radar,yang2025ground,nissov2024degradation}, Doppler radars~\cite{barlow1949doppler}, advanced 4D automotive radars~\cite{yang2025ground,zhuang20234d}, and 4D imaging radars~\cite{zhang20234dradarslam}. Each category exhibits distinct trade-offs in resolution, point cloud density, and robustness, which critically affect perception and state estimation in complex environments. Among these, 4D radars~\cite{zhuang20234d},~\cite{zhang20234dradarslam} often produce rich point cloud structures, facilitating pose estimation through scan matching algorithms like ICP~\cite{zhang20234dradarslam}. The relative pose obtained from scan-matching, together with ego velocity estimates and IMU measurements, is fused within an Extended Kalman Filter (EKF) to generate odometry. However, millimetre-wave~(FMCW TI IWR6843AOP EVM) radars inherently produce sparse point clouds with limited structural features, which significantly degrade the performance of scan-matching algorithms.

In contrast, prior work on EKF-RIO~\cite{doer2020ekf} and its online calibration~\cite{kim2025ekf} demonstrated that a single EKF using 4D radar can effectively estimate pose directly from radar point clouds. Our field experiments with a mobile robotic platform, equipped with two different types of IMU sensors and continuous FMCW radars, revealed that the state-of-the-art RIO approach becomes unreliable when the IMU is affected by dynamic biases, such as temperature-dependent sensor drift or uncompensated forward components of gravitational acceleration. Motivated by these observations and the limitations of single-radar EKF methods under IMU biases, we propose the Multi-Radar Inertial Odometry (MRIO) framework. MRIO introduces a two-stage EKF fusion strategy that integrates multiple FMCW radars with inertial measurements, enabling robust localization and mapping in subterranean, GPS-denied, and visually degraded environments.
\subsection{Contribution}
\begin{figure*} 
\centering
\includegraphics [width=1.0\textwidth]{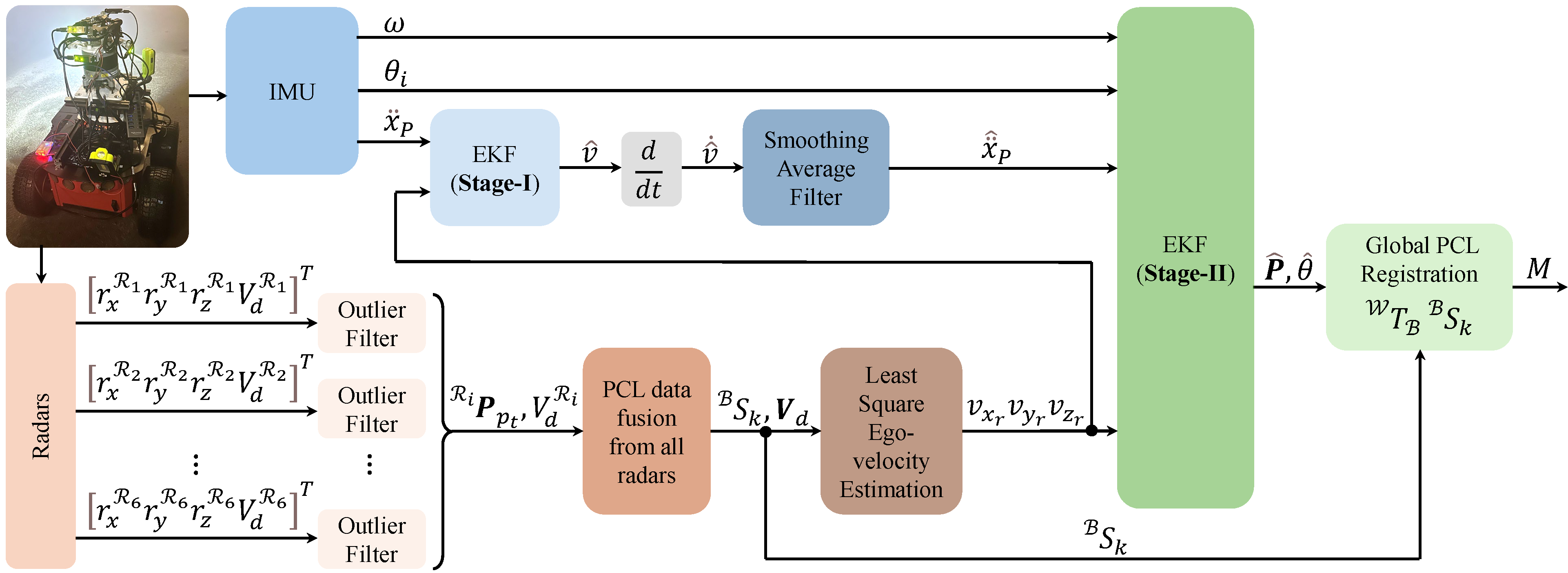}
\setlength{\belowcaptionskip}{-5pt}
\caption{\justifying A schematic of the Multi-Radar Inertial Odometry (MRIO) framework, which integrates inputs from an IMU and multiple radars mounted on a mobile robotic platform to obtain odometry, ego velocity, and a radar point-cloud map. Radar scans, containing target points and Doppler velocities, are processed via a least-squares filter to obtain ego velocity, which is then refined using a two-stage filtering approach for accurate localization and mapping.}
\label{fig: MRio framework}
\end{figure*}

Unlike conventional methods, the proposed framework is designed for resilient navigation in harsh subterranean environments, withstanding extreme cold, dark, humid and sloped tunnels. A key focus is the role of IMU quality: while commonly used IMU like Pixhawk are exposed to dynamic biases that become unsuitable for state-of-the-art EKF-RIO, our two-stage MRIO design enables reliable radar–inertial odometry even under such conditions. This analysis highlights the impact of IMU characteristics on radar inertial fusion and identifies key considerations for implementing robust subterranean navigation. Figs.~\ref{fig: MRio framework},~\ref{fig:Cover_fig_1} illustrate our approach and its outcomes in comparison with the state-of-the-art EKF-RIO~\cite{kim2025ekf} and LIO-SLAM respectively. The key contributions are as follows:
\begin{itemize}
\item MRIO is, to the best of our knowledge, the first framework to demonstrate radar inertial odometry and mapping using multiple cost-effective IWR6843AOP EVM mmWave radars (each with a $60^{\circ}$ FOV, providing enhanced coverage). 
\item A sensing-radius outlier rejection method further outperforms RANSAC on sparse radar data, improving map quality across diverse environments.

\item In addition, the framework incorporates a complementary bias estimator that enables robust operation even with both low-cost and high-grade IMUs (e.g., Pixhawk and VectorNav), effectively addressing dynamic biases that conventional EKF-RIO cannot handle.

\item The framework is further validated online in subterranean tunnels over long-range distances and clear sloped passages, where its performance is benchmarked against EKF-RIO~\cite{kim2025ekf} and LiDAR-inertial odometry (LIO). In both evaluation scenarios, MRIO achieves robust localization and mapping, demonstrating resilience under visually degraded conditions.
\end{itemize}
\section{Methodology} \label{Methodology}
\subsection{Notation}
Reference frames are denoted by calligraphic letters, e.g., $\mathcal{W}$ for the world frame and $\mathcal{B}$ for the robot body frame. The rigid-body transformation from $\mathcal{B}$ to $\mathcal{W}$ is defined as
\begin{equation}
{}^{\mathcal{W}}\bm{T}_{\mathcal{B}}
= \begin{bmatrix}
{}^{\mathcal{W}}\bm{R}_{\mathcal{B}} & {}^{\mathcal{W}}\bm{P}_{\mathcal{B}} \\ 
\bm{0}^\top & 1
\end{bmatrix} \in SE(3),
\end{equation}
where ${}^{\mathcal{W}}\bm{R}_{\mathcal{B}} \in SO(3)$ is the rotation matrix and ${}^{\mathcal{W}}\bm{P}_{\mathcal{B}} \in \mathbb{R}^3$ is the translation vector.  

The notation ${}^{\mathcal{F}}\bm{P}_{p}$ denotes the coordinates of a point $p$ expressed in frame $\mathcal{F}$. A point represented in $\mathcal{B}$ can be transformed into $\mathcal{W}$ as
\begin{equation}
\begin{bmatrix}
{}^{\mathcal{W}}\bm{P}_{p} \\
1
\end{bmatrix}
=
{}^{\mathcal{W}}\bm{T}_{\mathcal{B}}
\begin{bmatrix}
{}^{\mathcal{B}}\bm{P}_{p} \\
1
\end{bmatrix}.
\end{equation}
The symbol $\bm{0}_{3\times 1}$ denotes a $3 \times 1$ zero vector, and the superscript $(\cdot)^\top$ indicates matrix transpose unless otherwise specified.The calligraphic symbol $\mathscr{X}$ denotes state vectors, while the letters in calligraphic font (e.g., $\mathcal{W}$ and $\mathcal{B}$) denote frames or structured sets, and $\mathcal{R}$ represents the sensing radius of each radar’s field of view (FOV).
\subsection{\textbf{Augmented Kinematics with Accelerometer Bias}}
The discrete-time kinematics of a non-holonomic mobile robot at time step $k$ is given by
\begin{equation}
\begin{aligned}
x_{k+1} &= x_k + v_k \cos\theta_k \, \Delta t, \\
y_{k+1} &= y_k + v_k \sin\theta_k \, \Delta t, \\
\theta_{k+1} &= \theta_k + \omega_k \, \Delta t,
\end{aligned}
\label{eq:classical_kinematics}
\end{equation}
where $[x_k, y_k]^\top$ denotes the robot’s position in the world frame, $\theta_k$ the orientation, $v_k$ the forward velocity in the body frame, and $\omega_k$ the angular velocity.  
\subsubsection{Augmented Dynamics}
In inertial navigation, unmodeled accelerometer biases $b_{a,k}$ lead to significant drift in velocity and position estimates. To account for this, the velocity dynamics are augmented as
\begin{equation}
\begin{aligned}
v_{k+1} &= v_k + \big(a_{c,k} - b_{a,k}\big)\Delta t, \\
b_{a,k+1} &= b_{a,k} + w_{b,k}\Delta t,
\end{aligned}
\label{eq:augmented_kinematics}
\end{equation}
where $a_{c,k}$ is the commanded acceleration and $w_{b,k}$ models the bias evolution. Two simplified cases are typically considered: 
\begin{itemize}
    \item \textbf{Case 1 (Constant Bias)}:  
    Setting $w_{b,k}=0$ yields
    \[
    b_{a,k+1} = b_{a,k},
    \]
    i.e., $b_{a,k}=c$ for all $k$. This enables bias calibration at initialization but ignores subsequent variations.  

    \item \textbf{Case 2 (Linearly Time-Varying Bias)}:  
    Setting $w_{b,k}=c$ (a constant drift) yields
    \[
    b_{a,k+1} = b_{a,k} + c\Delta t \quad \Rightarrow \quad b_{a,k}=b_{a,0}+k\,c\Delta t.
    \]
    The bias grows linearly with time, independent of the actual acceleration, and may diverge arbitrarily.  
\end{itemize}
\subsubsection{Limitations of Augmented Kinematics}Both cases provide limited representations of accelerometer bias:  
\begin{itemize}
    \item Case 1 is overly restrictive, failing to capture environmental effects such as temperature fluctuations or gravity projection errors.  
    \item Case 2 produces a deterministic, unbounded drift term that lacks correlation with the underlying dynamics.  
\end{itemize}
\subsection{\textbf{Stage-I: Bias-Aware EKF for Inertial Offset Compensation}}\label{ref: stage: I}
To overcome these limitations, we adopt an EKF-based offset estimation approach, where the IMU acceleration measurement is explicitly modelled as being affected by both bias and noise. Within the MRIO framework~(Fig.~\ref{fig: MRio framework}), this procedure constitutes Stage-I.
\begin{align}
\hat{v}_{p,{k+1}} &= v_{p,k}+(a_{c,{k}} + a_{c,{b_k}} + a_{n,{k}})\Delta t, 
\label{eq:offset_kinematics}
\end{align}
where $a_{c,{b_k}}$ denotes the offset and $a_{n,{k}}$ the additive noise. Unlike constant or deterministic drift models, this formulation does not constrain the bias to a predefined law but instead estimates it online within the EKF~\cite{simon2006optimal} framework. This enables robust compensation of IMU bias under the severe environmental variations characteristic of subterranean tunnel experiments.
\begin{figure}[h] 
\centering
\includegraphics [width=0.49\textwidth]{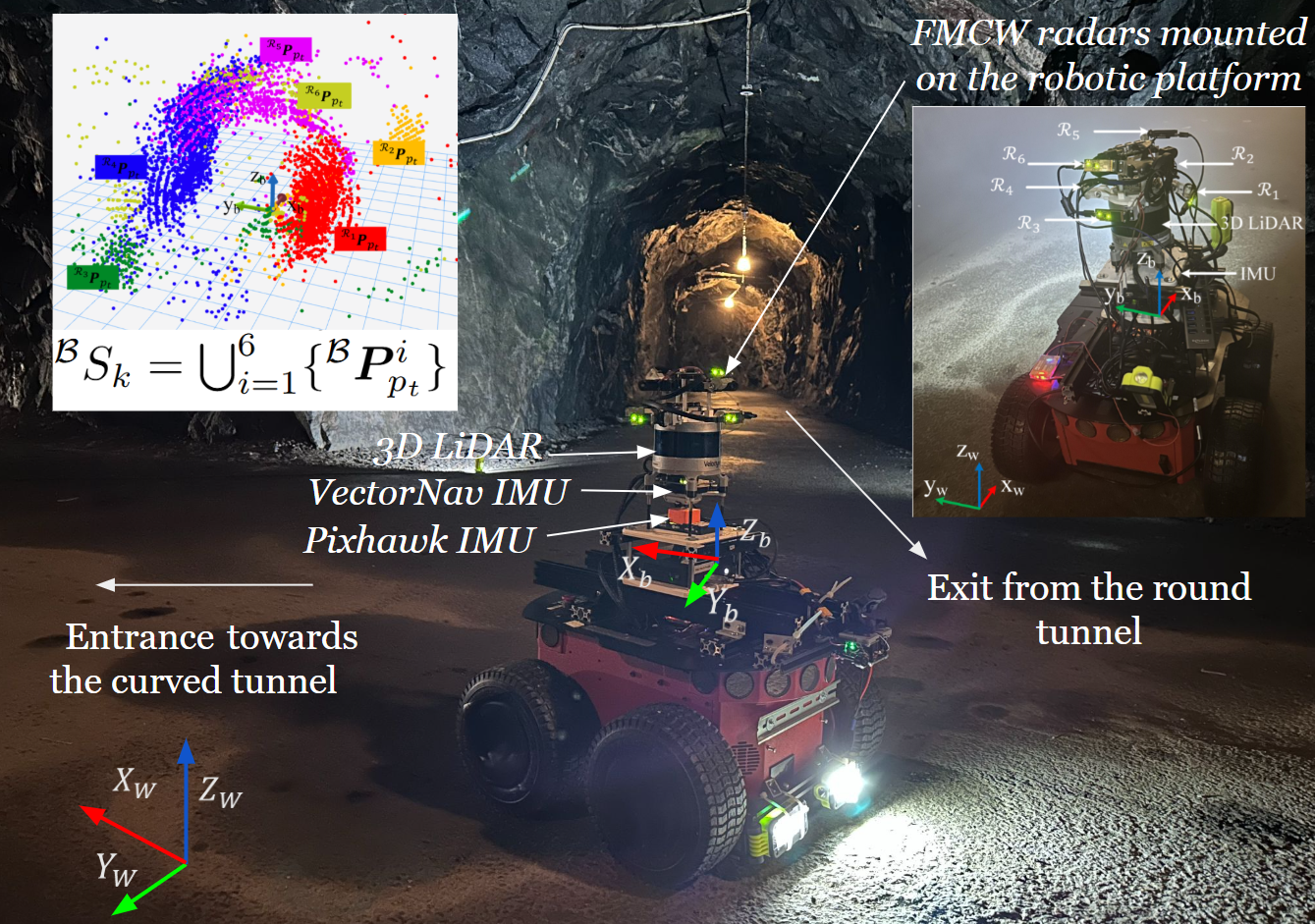}
\setlength{\belowcaptionskip}{-5pt}
\caption{\justifying \textbf{On the left:} The experimental setup includes six mmWave TI-IWR6843AoP compact radars (labeled $R_1$ to $R_6$), arranged in an ensemble configuration with LiDAR and IMU, mounted on the Pioneer 3-AT2 rover platform. Here, radars $R_1-R_4$ are mounted on the same plane, while radars $R_5-R_6$ are positioned at a slanted angle, capturing the perception of the ceiling from both the front and rear sides. \textbf{On the right:} Visualization of radar point clouds captured at a single moment using the proposed multi-radar setup. The target points are colour-coded, with each colour representing different radar target points${}^{\mathcal{R}_i}_{}\bm{P}_{p_t} \forall i= 1\dots 6$}
\label{fig: sensor_setup}
\end{figure}
Unlike constant or deterministic drift models, the proposed Stage-I formulation does not constrain the accelerometer bias to a predefined law but instead estimates it online within an EKF framework with the help of estimated radar-based velocity. This enables robust compensation of IMU bias under the severe environmental variations characteristic of a subterranean tunnel.

In parallel, radar measurements ${}^{\mathcal{B}}\bm{P}_{p_c}$ and their associated Doppler velocities $\hat{v}^d$ are used to obtain an initial ego-velocity estimate $v_r$ through least-squares fitting. Each radar $i \in \{1,\dots,6\}$ provides point clouds ${}^{\mathcal{R}_i}\bm{P}_{p_t}=[r^{R_i}_x,r^{R_i}_y,r^{R_i}_z]^\top$ in its local frame, which are filtered within the feasible sensing radius $\mathcal{R}_I \leq \mathcal{R}_e \leq \mathcal{R}_O$ and transformed into the body frame $\mathcal{B}$. The transformed scans from all radars are merged to form the radar point cloud ${}^\mathcal{B}S_k=\bigcup_{i=1}^{6}\{{}^{\mathcal{B}}\bm{P}_{p_t}^i\}$. Using the corresponding Doppler velocities $V_d^{R_i}$, the ego-velocity $\hat{v}^r$ is then obtained by solving the overdetermined system
\begin{align}
\label{Eq:Ls soln_final}
\hat{v}^r = (H^\top H)^{-1} H^\top y^r,
\end{align}
where $H$ encodes the unit vectors of target points and $y^r$ stacks the Doppler measurements. Since least squares radar velocities~\cite{kramer2020radar},\cite{kingery2022improving}~\cite{rai20234dego},~\cite{nissov2024roamer} are less accurate than their LiDAR-derived velocities, they are refined in Stage I using an EKF that fuses them with IMU acceleration, thereby compensating for bias and noise. The resulting ego-velocity provides an offset-free acceleration (via differentiation and smoothing), which is then passed to Stage-II. In Stage-II, corrected acceleration, radar velocity updates, and IMU orientation are fused within an EKF node to estimate the global pose and register radar point clouds. Thus, Stage-I stabilizes the inertial inputs, while Stage-II achieves accurate global localization and mapping.
\begin{algorithm}[t]
\footnotesize
\caption{MRIO: Multi-Radar Inertial Odometry with Bias-Aware IMU Drift Estimator}
\label{algo:MRIO_algorithm_multirate_online}
\begin{algorithmic}[1]
\Require IMU $\{t_k, \omega_k, \dot v_k\}$, Radar $\{t_m,\ R^{(r)}_{x,m},R^{(r)}_{y,m},R^{(r)}_{z,m},V^{(r)}_{d,m}\}_{r=1}^6$
\Ensure $\hat v_k,\ \hat a_{cc,k},\ \hat p_k,\ \hat\theta_k,\ M_m$
\State $\mathscr{X}_{0|0}\gets \mathscr{X}_0,\ \Sigma_{0|0}\gets\Sigma_0$
\State $t^- \gets \bot,\ \hat v^- \gets \bot,\ \hat a_{cc,0}\gets 0$
\For{$k=1$ to $K$}
  \State $(\mathscr{X}^-,\Sigma^-) \gets f_{\text{integrate}}(\mathscr{X}_{k-1},\Sigma_{k-1},\omega_k,\dot v_k,\Delta t_k)$
  \State $\hat p_k \gets \Pi_p(\mathscr{X}^-),\ \hat v_k \gets \Pi_v(\mathscr{X}^-),\ \hat\theta_k \gets \Pi_\theta(\mathscr{X}^-)$
  \If{radar update $t_m \in [t_{k-1},t_k]$}
    \ForAll{$m$ in $[t_{k-1},t_k]$ (asc.)}
      \State $\tilde{\mathcal Z}_m^{(r)} \gets \mathrm{Filt}(\{R^{(r)}_{x,y,z,m},V^{(r)}_{d,m}\})$
      \State ${}^\mathcal{B}S_m \gets \mathrm{Merge}(\tilde{\mathcal Z}_m^{(r)})$
      \State $\hat v_{\text{lsq}} \gets g(\tilde{\mathcal Z}_m^{(r)})$ \Comment{LSQ ego-velocity}
      \State $\hat v_I \gets f_{\text{stage-I}}(\mathbf{p}_{k-1},\dot v_k,\Delta t_k)$
      \State $\hat a_{cc,k}\gets (\hat v_I-\hat v^-)/(t_m-t^-)$ if $t^-\neq\bot$ else $0$
      \State $z_k \gets \hat v_{\text{lsq}}$
      \State $H_k \gets \partial h/\partial \mathscr{X}|_{\mathscr{X}^-},\quad \nu_k \gets z_k-h(\mathscr{X}^-)$
      \State $S_k \gets H_k \Sigma^- H_k^\top+R_m,\quad K_k \gets \Sigma^- H_k^\top S_k^{-1}$
      \State $\mathscr{X}_{k|k}\gets \mathscr{X}^-+K_k\nu_k,\quad$
      \State $\Sigma_{k|k}\gets (I-K_kH_k)\Sigma^-(I-K_kH_k)^\top+K_kR_mK_k^\top$
      \State $M_m \gets ({}^\mathcal{B}S_m,\ {}^\mathcal{W}T_{\mathcal{B}}(\hat p_k,\hat\theta_k))$
      \State $(t^-,\hat v^-) \gets (t_m,\hat v_{\text{lsq}})$
    \EndFor
  \EndIf
\EndFor
\end{algorithmic}
\end{algorithm}
\subsection{\textbf{Real-Time Radar Inertial Fusion for Localization and Map Registration(Stage-II)}}
\label{stage:II_RIO}
Stage-II~(Fig.~\ref{fig: MRio framework}) achieves real-time radar–inertial fusion by integrating high-rate IMU signals with lower-rate radar updates in an EKF framework. This is the node where drift-free linear acceleration and radar measurements are fused in a centralized node through EKF~\cite{doer2020ekf}. The bias-corrected acceleration from Stage-I is propagated at the IMU frequency, ensuring fine-grained state prediction, while radar-derived ego-velocity estimates at 10 Hz provide asynchronous measurement updates. This multi-rate design enables radar–inertial odometry (RIO) suitable for online global point cloud registration.  
\begin{figure*}[t]
    \centering
    \captionsetup{font=footnotesize,aboveskip=1pt,belowskip=1pt}

    \begin{subfigure}[t]{0.48\textwidth}
        \centering
        \includegraphics[width=\linewidth]{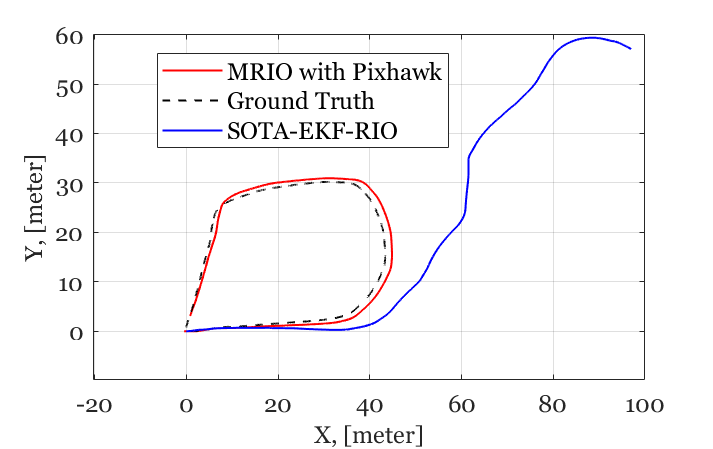}
        \caption{MRIO vs. EKF\textendash RIO with Pixhawk IMU.}
        \label{fig:trajectory_px4}
    \end{subfigure}
    \hfill
    \begin{subfigure}[t]{0.48\textwidth}
        \centering
        \includegraphics[width=\linewidth]{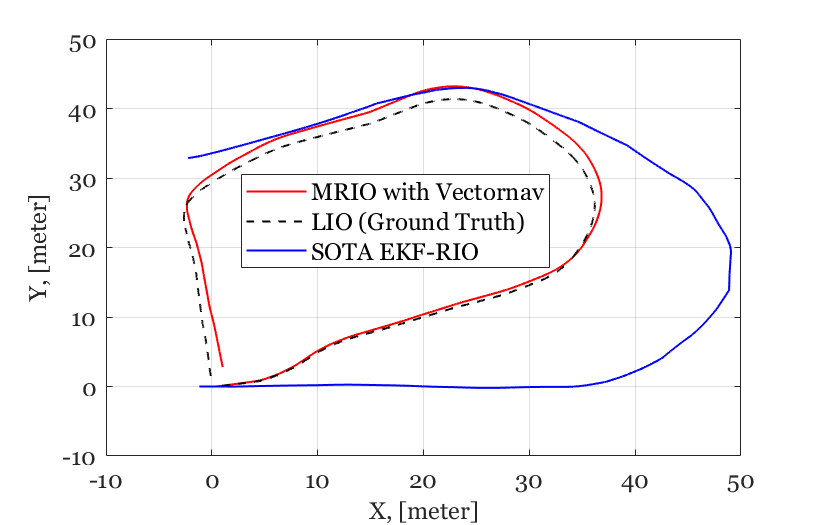}
        \caption{MRIO vs. EKF\textendash RIO with VectorNav IMU.}
        \label{fig:trajectory_vectornav}
    \end{subfigure}

    \caption{Localization performance comparison across IMUs using MRIO and EKF\textendash RIO.}
    \label{fig:mrio_sota_trajectories}
    \vspace{-2mm} 
\end{figure*}
The EKF estimates the state vector
\[
\mathscr{X}^{II}_{k}=
\begin{bmatrix}
x^{II}_k,\ y^{II}_k,\ \hat{v}_{p,k},\ \theta_k
\end{bmatrix}^\top,
\]
representing the robot’s position, velocity, and orientation. The innovation vector compares radar and IMU measurements against predicted states:
\begin{align}
\nu_{m,k} &= z^{II}_{m,k} - h(\hat{\mathscr{X}}^{II}_{m,k|k-1}),
\end{align}
where $z^{II}_{k} = [\theta_k,\hat{v}_{r,k}]^\top$ consists of the ego velocity estimated from radar least-squares estimation (Eq.~\ref{Eq:Ls soln_final}) and the IMU orientation.
where error covariance,
\begin{align}
J(\tilde{\mathcal{X}}) = \tfrac{1}{2} \tilde{\mathcal{X}}_k^\top S_k^{-1} \tilde{\mathcal{X}}_k,
\end{align}
where,
\begin{align}
S_k = H_{k}\Sigma^{-}_{k}H^T_k+R_k
\end{align}
Minimizing error covariance $J(\tilde{\mathcal{X}})$ yields the state estimate that best balances the process model and measurement uncertainties. The Kalman gain:
$K_k = \Sigma_{k|k-1} H_k^\top S_k^{-1}$
where $\Sigma_{k|k-1}$ is the predicted state covariance and $H_k$ the Jacobian of $h(\cdot)$, determines the optimal weighting between prediction and correction. Compared to Stage-I, where bias compensation dominates, the innovation vector in Stage-II directly compares the velocity propagated from offset-free acceleration with radar-based velocity estimates, resulting in reduced error. This makes Stage-II EKF essential for stable global localization, especially under the adverse conditions of subterranean environments.  The estimated pose $(\hat{P}, \hat{\theta})$ provides the translational and rotational components of the homogeneous transformation matrix
\[
{}^\mathcal{W}T_{\mathcal{B}} =
\begin{bmatrix}
R & P \\
0^\top & 1
\end{bmatrix},
\]
which converts radar point clouds from the robot’s body frame $\mathcal{B}$ to the world frame $\mathcal{W}$. At each time step, transformed scans are accumulated to form the global map.
\begin{equation}
M = {}^\mathcal{W}T_{\mathcal{B}}\, {}^\mathcal{B}S_k,
\end{equation}
where ${}^\mathcal{B}S_k$ denotes the radar scan in the body frame. This process incrementally constructs a consistent global map.
\begin{table}[h]
\footnotesize
\centering
\caption{Configuration Parameters for 3D-Velodyne LiDAR}
\setlength{\tabcolsep}{2.25mm}
\renewcommand{\arraystretch}{1.2} 
\begin{tabular}{|c|c|c|c|c|}
\hline
\centering
\textbf{Parameter} & \textbf{Value}\\ \hline
\centering
\textbf{Laser Channels} &16  \\ \hline
\textbf{Vertical Resolution} &0.2° \\ \hline
\textbf{Field of View (FOV)} & 360°  \\ \hline
\textbf{Range} & 100m  \\ \hline
\textbf{Rotation Rate} &(1-100) Hz\\ \hline
\textbf{Points per second} &~300,000 points/sec  \\ \hline
\end{tabular}
\label{Table: 3D velodyne Lidar}
\end{table}
\begin{table}[h]
\footnotesize
\centering
\caption{Configuration Parameters for Millimeter-wave IWR6843AOP EVM mmWave radars}
\setlength{\tabcolsep}{2.25mm}
\renewcommand{\arraystretch}{1.2} 
\begin{tabular}{|c|c|c|c|c|}
\hline
\centering
\textbf{Parameter} & \textbf{Value}\\ \hline
\centering
\textbf{Frame-rate} & 8.64-10 GHz\\ \hline
\textbf{Band width} & 60-64Hz \\ \hline
\textbf{Maximum range} & 8-10 meters   \\ \hline
\textbf{Maximum doppler} &  8-10m/s \\ \hline
\textbf{Range resolution} &0.047  \\ \hline
\textbf{Doppler velocity resolution} &3 cm/s   \\ \hline
\end{tabular}
\label{Table: radar_specifications}
\end{table}
\section{Experimental Evaluation}
\begin{table*}[t]
\footnotesize
\centering
\caption{\textbf{IMU Specifications} (a) PX4 Cube Orange; (b) VectorNav VN-100. 
Symbols: \(\sigma_\omega\) gyro noise density, \(\beta_\omega\) gyro bias stability, 
\(\sigma_a\) accel noise density, \(\beta_a\) accel bias stability.}
\setlength{\tabcolsep}{4mm}
\renewcommand{\arraystretch}{1.2}

\begin{subtable}[t]{0.47\textwidth}
\centering
\caption{PX4 Cube Orange}\label{Tab: Px-4 IMU}
\begin{tabular}{|c|c|}
\hline
\textbf{Parameter} & \textbf{Value} \\ \hline
Manufacturer & CubePilot Pvt.\ Ltd \\ \hline
Model & Cube Orange \\ \hline
Gyro range & \(\pm 250 \ldots \pm 2000~\mathrm{dps}\) \\ \hline
\(\sigma_\omega\) (gyro) & \(4~\mathrm{mdps}/\sqrt{\mathrm{Hz}}\) \\ \hline
\(\boldsymbol{\beta_\omega}\) (gyro) & \textit{not specified} \\ \hline
Accel range & \(\pm 2g \ldots \pm 16g\) \\ \hline
\(\sigma_a\) (accel) & \(100~\mu g/\sqrt{\mathrm{Hz}}\) \\ \hline
\(\boldsymbol{\beta_a}\) (accel) & \textit{not specified} \\ \hline
Update rate & up to \(\sim 1~\mathrm{kHz}\) (firmware dep.) \\ \hline
Weight / size & \(35~\mathrm{g}\) / \(38.4\times 38.4\times 22~\mathrm{mm}^3\) \\ \hline
\end{tabular}
\end{subtable}
\hfill
\begin{subtable}[t]{0.47\textwidth}
\centering
\caption{VectorNav VN-100}\label{Tab: Vectornav}
\begin{tabular}{|c|c|}
\hline
\textbf{Parameter} & \textbf{Value} \\ \hline
Manufacturer & VectorNav \\ \hline
Model & VN-100 IMU/AHRS \\ \hline
\(\boldsymbol{\beta_\omega}\) (gyro) & \(< 3.5~\deg/\mathrm{hr}\) (typ.) \\ \hline
\(\sigma_\omega\) (gyro) & \(\approx 0.0035~\deg/\sqrt{\mathrm{s}}\) \\ \hline
\(\boldsymbol{\beta_a}\) (accel) & \(< 0.04~\mathrm{mg}\) \\ \hline
\(\sigma_a\) (accel) & \(\approx 0.14~\mathrm{mg}/\sqrt{\mathrm{Hz}}\) \\ \hline
Onboard EKF update & \(400~\mathrm{Hz}\) \\ \hline
Raw data rate & \(800~\mathrm{Hz}\) \\ \hline
Pitch/Roll accuracy & \(0.5^{\circ}\) \\ \hline
Weight / size & \(15~\mathrm{g}\) / \(24\times 22\times 6~\mathrm{mm}^3\) \\ \hline
\end{tabular}
\end{subtable}
\end{table*}
Field experiments were conducted using multiple FMCW radars and two IMU configurations mounted on a ground robot to evaluate the performance of the proposed multi-radar inertial odometry (MRIO) framework. Its performance was compared against the state-of-the-art EKF-RIO, equipped with online calibration, as well as a LiDAR-based SLAM approach. The experiments were conducted in subterranean tunnels, where ground-truth references such as motion capture or GPS were unavailable. These tunnels posed severe challenges, characterized by cold, humid, and dark conditions, with irregular wall surfaces and slopes. In such environments, IMUs mounted on ground robots are prone to biases induced by temperature variations and gravitational components during motion inside the cave.
\begin{figure}[h] 
    \centering
    \begin{subfigure}[t]{\linewidth}
        \centering
        \includegraphics[width=1.1\linewidth]{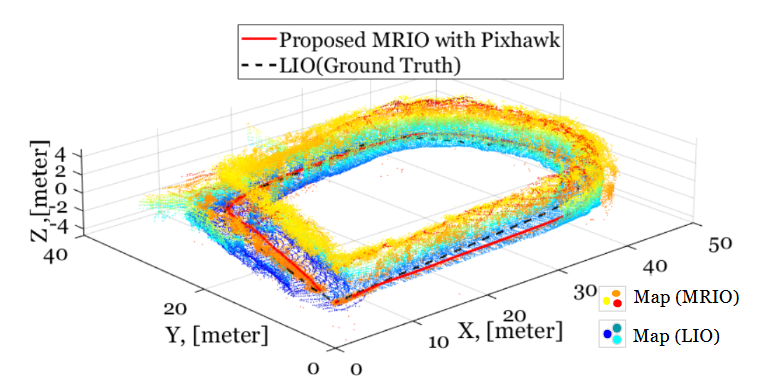}
        \caption{PX4 IMU.}
        \label{fig:mrio_px4}
    \end{subfigure}
\vspace{0.5em}
    \begin{subfigure}[t]{\linewidth}
        \centering
        \includegraphics[width=1.1\linewidth]{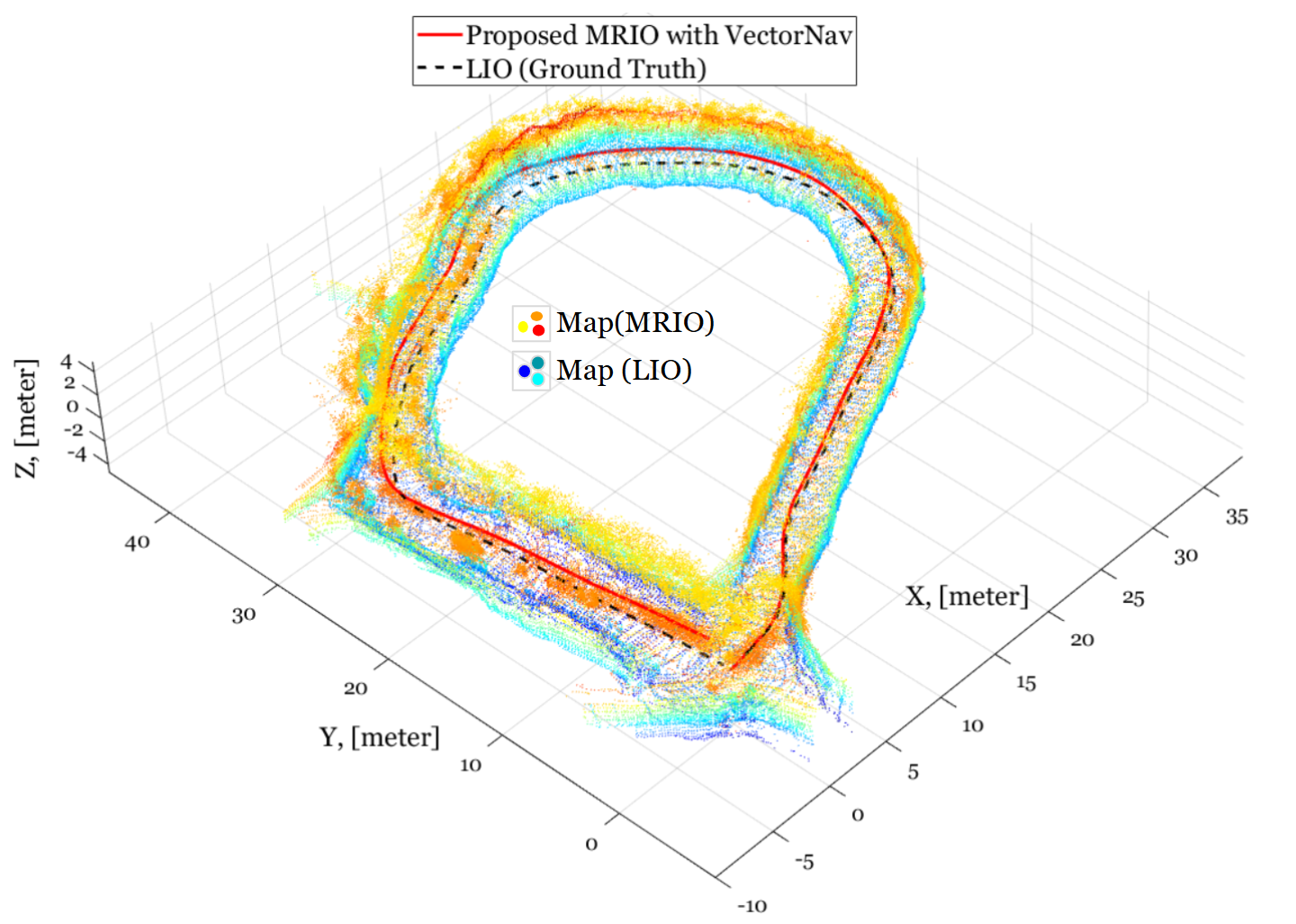}
        \caption{VectorNav IMU.}
        \label{fig:mrio_vectornav}
    \end{subfigure}
    \caption{\justifying Experiments in the SubT tunnel with PX4 and VectorNav IMUs demonstrate that the proposed framework achieves robust localization and mapping performance compared to LIO-SLAM across all scenarios.}
    \label{fig:rio_acceleration_comparison}
\end{figure}
To address this limitation, we propose a multi-stage RIO with a dedicated bias estimator, evaluated on two IMU classes: (i) units with built-in bias estimation (Table~\ref{Tab: Px-4 IMU}) and (ii) cost-effective units lacking hardware bias estimation (Pixhawk 2.1 Cube Orange: Table~\ref{Tab: Vectornav}). This design improves robustness over EKF-RIO with online calibration~\cite{kim2025ekf} for radars, which we have used. We arrange multiple $60^{\circ}\times60^{\circ}$ radars for near-$360^{\circ}$ body-frame coverage, the resulting point clouds are sparse, noisy, and multipath-cluttered (Fig.~\ref{fig: sensor_setup}), making conventional scan matching ineffective. Sec .~\ref {sec: Experimental Setup} describes the specifications of radars, 3D Velodyne LiDAR, and IMU sensors used in our experiments.
\begin{figure}[h]
    \centering
    \begin{subfigure}[t]{0.35\textwidth}
        \centering
        \includegraphics[width=\linewidth]{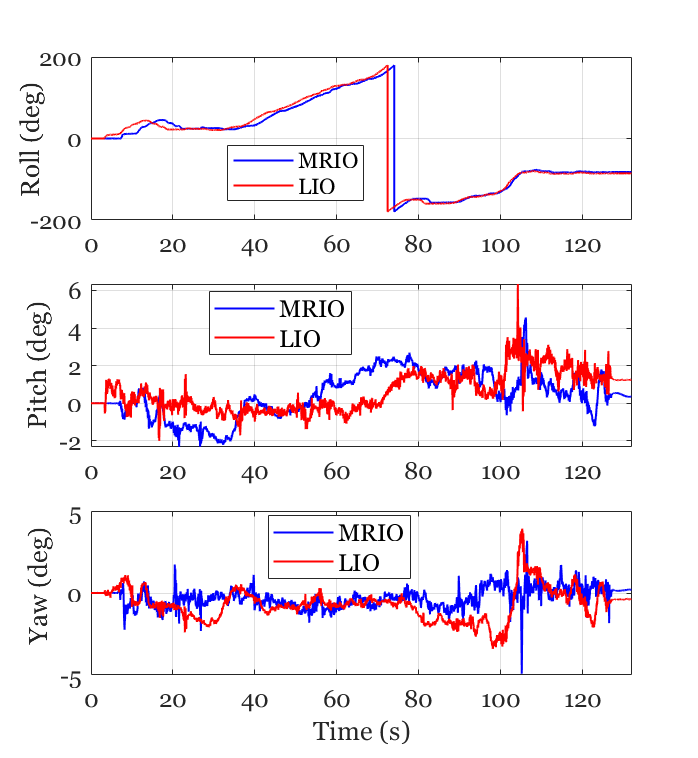}
        \caption{PX4 IMU.}
        \label{fig:ori_mrio_px4}
    \end{subfigure}
    \hfill
    \begin{subfigure}[t]{0.35\textwidth}
        \centering
        \includegraphics[width=\linewidth]{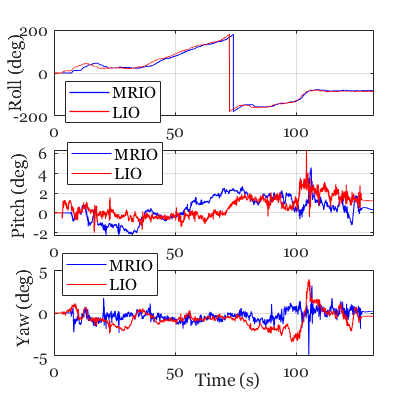}
        \caption{VectorNav IMU.}
        \label{fig:ori_mrio_vectornav}
    \end{subfigure}
    \caption{\justifying Estimated orientation from the proposed algorithm with LIO SLAM for Pixhawk and VectorNav IMU.}
    \label{fig:mrio_orientations}
\end{figure}
\subsection{Experimental Setup}\label{sec: Experimental Setup}
In this study, we employ the Pioneer 3-AT platform, which is equipped with IWR6843AOP EVM millimetre-wave radars in an ensembled configuration, enabling perception with enhanced field of view~\cite{vales2024iot} and with two types of IMUs (VectorNav~\cite{hartzer2023online} and Pixhawk~\cite{corrigan2025design}). In this hardware setup, the radar sensors operate within the frequency range of 60 GHz to 64 GHz. They are strategically placed at $90$-degree angles relative to each other, set to face forward, left, backwards and right, as depicted in Fig.~\ref{fig: sensor_setup}. Two more radars, positioned in a slightly tilted manner relative to the $2D$ plane, are integrated into the setup. These radars are oriented at a $45$-degree angle and can capture PCLs above the vehicle's front and rear. This setup improves the area covered by the six radar units more effectively than a configuration with fewer radars. Since all experiments are conducted in subterranean environments, a 3D Velodyne LiDAR-based SLAM is used as the ground truth when the environment is free from smoke. The tables below summarize the configurations of the radars~(Table~\ref{Table: radar_specifications}), 3D Velodyne LiDAR~(Table~\ref{Table: 3D velodyne Lidar}), and IMUs~(Table~\ref{Tab: Px-4 IMU},~\ref{Tab: Vectornav}). For online implementation, we evaluated EKF-RIO with two IMU configurations to assess the impact of drift accumulation. 
\begin{figure*}[h]
    \centering
    \captionsetup{font=footnotesize,aboveskip=2pt,belowskip=2pt}
    \begin{subfigure}[t]{0.45\linewidth}
        \centering
        \includegraphics[width=\linewidth]{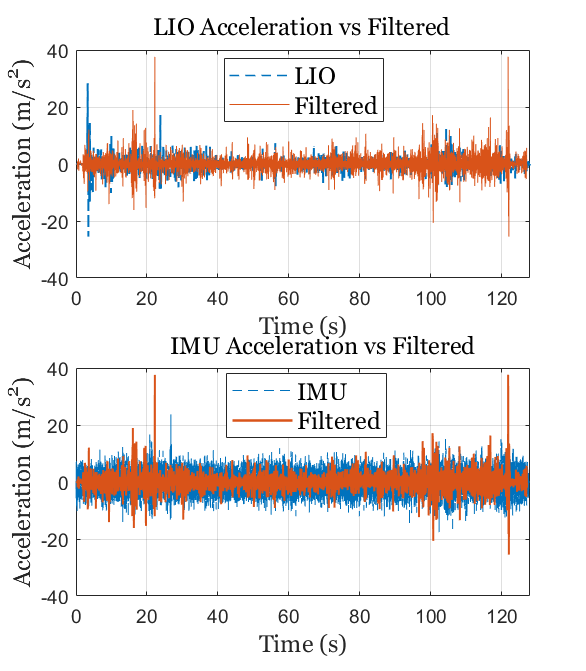}
        \caption{PX4 IMU}
        \label{fig:mrio_acc_correction_px4}
    \end{subfigure}\hfill
    \begin{subfigure}[t]{0.49\linewidth}
        \centering
        \includegraphics[width=\linewidth]{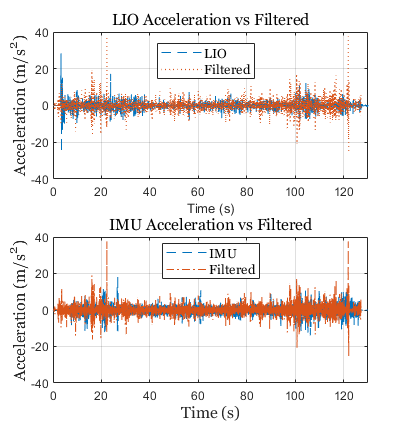}
        \caption{VectorNav IMU}
        \label{fig:mrio_acc_correction_vectornav}
    \end{subfigure}

    \caption{Acceleration (x) comparison for PX4 and VectorNav IMUs.}
    \label{fig:rio_acceleration_comparison}
    \vspace{-2mm} 
\end{figure*}
\subsection{MRIO Implementation in Subterranean Environments}
To implement and evaluate our framework, we conducted experiments in a tunnel environment with a 100-meter rounded trajectory, as illustrated in Fig.~\ref{fig: sensor_setup}, which serves as a benchmark for comparing our method with the state-of-the-art EKF-RIO employing online calibration on both Pixhawk and VectorNav. During the development of the framework and experimental setup, we observed that pedestrian movements and other dynamic activities within the environment significantly influenced ego-velocity estimation~\cite{chen2024robust},~\cite{chen2023drio}~\cite{nissov2024robust}, thereby degrading overall odometry performance. To address these challenges and ensure reliable monitoring and control in these tunnels, we employed Walksnail Avatar HD FPV goggles, which provided a direct visual feedback channel for navigating the robot. Under these conditions, we evaluated the performance of EKF-RIO with online calibration and our proposed framework using both Pixhawk and VectorNav sensors. 

In general these radar sensors operate based on relative velocity measurements, and they do not produce sufficient returns under static conditions. Once the robot begins moving, however, the radar package captures raw target points reflected from environmental structures such as walls, the ground, and the ceiling, together with Doppler velocities induced by relative motion. In practice, mmWave radars are also susceptible to multi-path propagation, which often leads to dense clusters of spurious target points accumulating in the vicinity of the robot. To mitigate these effects, we replace RANSAC~\cite{quist2016radar} with a range-based outlier rejection scheme, as RANSAC performed poorly on our sparse, flickering radar point clouds. This method exploits the expected detection range characteristics of the experimental environment and is applied to each time-stamped radar scan, thereby filtering erroneous measurements prior to subsequent ego-velocity estimation(sec.~\ref{Eq:Ls soln_final}) for the two experimental setups. These ego-velocity estimates are then fused with IMU measurements for acceleration correction within an EKF node, referred to as Stage-I filtering (sec.~\ref{ref: stage: I}). The outcomes of radar-assisted acceleration correction are illustrated in Fig.~\ref{fig:mrio_acc_correction_px4} and Fig.~\ref{fig:mrio_acc_correction_vectornav}, for the Pixhawk and VectorNav IMUs, respectively, also with LIO-derived acceleration.
\begin{table*}[h]
\centering
\caption{Quantitative comparison of MRIO (Pixhawk, VectorNav) and SOTA EKF-RIO against the ground truth. The lowest errors are highlighted in \textbf{bold}.}
\label{tab:errors}
\begin{tabular}{|l|c|c|c|c|}
\hline
Method & 2D RMSE [m] & RMSE X [m] & RMSE Y [m] & RMSE Yaw [$^\circ$] \\
\hline
MRIO (Pixhawk)          & \textbf{0.83} & \textbf{0.52} & \textbf{0.73} & \textbf{1.13} \\
SOTA EKF-RIO (Pixhawk)  & 42.72          & 38.21         & 19.10         & 2.01 \\
\hline
MRIO (VectorNav)        & \textbf{0.81} & \textbf{0.58} & \textbf{0.63} & \textbf{1.13} \\
SOTA EKF-RIO (VectorNav)& 38.60          & 30.68         & 23.42         & 1.13 \\
\hline
\end{tabular}
\end{table*}
Fig.~\ref{fig:mrio_px4} illustrates the implementation of the proposed MRIO framework using a Pixhawk IMU, which does not incorporate onboard bias estimation. In the cold and inclined subterranean tunnel environment, temperature-induced effects further exacerbated sensor imperfections, leading to noticeable degradation in the IMU measurements, particularly in the linear acceleration. The accumulated inaccuracies rapidly propagated, leading to significant drift in the radar–inertial fusion process. This behaviour highlights a key limitation of single-stage fusion with low-cost IMUs, such as the Pixhawk, where uncompensated biases and temperature sensitivity lead to unreliable long-term odometry. As illustrated in Fig.~\ref{fig:trajectory_px4}, the state-of-the-art framework exhibits immediate drift and ultimately collapses into dead-reckoning during prolonged tunnel traversals. In contrast, the proposed two-stage Multi-Radar Inertial Odometry (MRIO) framework effectively addresses these limitations by exploiting redundant radar measurements to correct linear acceleration bias dynamically. Consequently, the estimated trajectories produced by the MRIO framework remain closely aligned with the ground-truth SLAM reference.

On the other hand, A similar evaluation with the VectorNav IMU(Fig.~\ref{fig:mrio_vectornav}), which consists of in-built advanced onboard bias estimator and temperature compensator, revealed that the single-stage EKF-RIO achieved better performance than with the Pixhawk, illustrated in Fig.~\ref{fig:trajectory_vectornav}, where odometry from state of the art framework follows the ground truth more closely compared to the performance with Pixhawk. Across both IMU configurations, MRIO consistently outperformed the SOTA algorithm~\cite{kim2025ekf}. These results demonstrate MRIO’s adaptability across IMU grades, especially on ground robots in subterranean tunnels. In our experiments, we observed no appreciable bias across the gyroscope axes; thus, orientation remained stable, illustrated in Figs.~\ref{fig:ori_mrio_px4} and~\ref{fig:ori_mrio_vectornav} and did not affect radar inertial odometry.
\subsection{Quantitative Analysis}
In terms of global accuracy, the proposed MRIO framework with Pixhawk (Fig.~\ref{fig:trajectory_px4}) achieved a 2D RMSE~(Table.~\ref{tab:errors})  of 0.83~m, significantly outperforming the baseline SOTA radar inertial odometry at 42.72~m (38.21~m in $x$, 19.10~m in $y$), while also reducing yaw RMSE (1.13$^\circ$ vs.~2.01$^\circ$). Similarly, in the VectorNav comparison (Fig.~\ref{fig:trajectory_vectornav}), MRIO closely followed the LiDAR--inertial ground truth, attaining a 2D RMSE of 0.81~m compared to 38.60~m for SOTA, corresponding to a $\sim$95\% improvement, with error reductions of $\sim$98\% in $x$ and $\sim$93\% in $y$. In both cases, MRIO maintained a consistently low relative translation error of $\sim$0.006~m/m (0.6\%). This underscores the robustness of the proposed framework in local pose estimation while simultaneously delivering substantially higher global trajectory fidelity.

\section{Conclusion}
We presented a multi-radar inertial odometry (MRIO) framework for resilient navigation in subterranean environments where LiDAR/vision degrades. MRIO adopts a two-stage design: radar ego-velocity (from cost-effective FMCW sensors) is estimated via least squares and injected into an EKF for online IMU bias correction, after which corrected inertial data are fused with multi-radar measurements. The framework supports radar-only mapping via estimated translational and rotational displacements and employs a sensing-radius outlier rejection scheme that outperforms RANSAC on sparse radar returns. Extensive tunnel trials, long traversals, slopes, and visually degraded conditions show consistent gains over state-of-the-art EKF-RIO and LIO across IMU grades. With a Pixhawk, MRIO achieved a 2D RMSE of 0.83\,m versus 42.72\,m for the baseline (38.21\,m in \(x\), 19.10\,m in \(y\)) and reduced yaw RMSE to \(1.13^\circ\) from \(2.01^\circ\) with a VectorNav, MRIO attained 0.81\,m versus 38.60\,m (\(\sim95\%\) improvement; \(\sim98\%\) in \(x\), \(\sim93\%\) in \(y\)) while maintaining a low relative translation error (\(\sim0.006\) m). These results highlight both the robustness of compact, lightweight, cost-effective radar–IMU setups and the critical influence of IMU quality in extreme cold and sloped tunnels.

A current limitation of the framework is its reliance on the IMU as the sole orientation source, which makes it susceptible to gyro-bias-induced drift and degraded map consistency. Future work will integrate redundant orientation cues and cross-sensor constraints to bound gyroscope bias, thereby improving robustness and global mapping fidelity in long-term subterranean deployments.
\bibliographystyle{IEEEtran}
\bibliography{Mou_bibliography.bib}
\end{document}